\documentclass{article}
\usepackage{spconf,amsmath,graphicx}
\usepackage{booktabs}
\usepackage{multirow}
\usepackage{url}
\usepackage[font=footnotesize]{subfig}
\usepackage{amssymb}
\usepackage{array}
\usepackage{makecell}
\usepackage{pifont} 
\usepackage{xcolor}   
\usepackage{float}
\usepackage{hyperref}

\newcommand{\symb}[2]{\raisebox{-0.15em}{\textcolor{#1}{\Large\ding{#2}}}}

\title{NEW VVC PROFILES TARGETING FEATURE CODING FOR MACHINES}
\name{
Md Eimran Hossain Eimon, Ashan Perera, Juan Merlos,  Velibor Adzic, Hari Kalva\vspace{-1ex}
}
\address{
Florida Atlantic University, USA
}

\begin{document}
%
\maketitle
\begin{abstract}
Modern video codecs have been extensively optimized to preserve perceptual quality, leveraging models of the human visual system. However, in split inference systems—where intermediate features from neural network are transmitted instead of pixel data—these assumptions no longer apply. Intermediate features are abstract, sparse, and task-specific, making perceptual fidelity irrelevant. In this paper, we investigate the use of Versatile Video Coding (VVC) for compressing such features under the MPEG-AI Feature Coding for Machines (FCM) standard. We perform a tool-level analysis to understand the impact of individual coding components on compression efficiency and downstream vision task accuracy. Based on these insights, we propose three lightweight essential VVC profiles--\textbf{Fast}, \textbf{Faster}, and \textbf{Fastest}. The Fast profile provides 2.96\% BD-Rate gain while reducing encoding time by 21.8\%. Faster achieves a 1.85\% BD-Rate gain with a 51.5\% speedup. Fastest reduces encoding time by 95.6\% with only a 1.71\% loss in BD-Rate.
\end{abstract}
\begin{keywords}
Feature coding, split inference, collaborative intelligence, coding for machines, VVC essential tools
\end{keywords}
\section{Introduction}
\label{sec:intro}
A large number of edge devices are capable of capturing visual data from cameras for computer vision (CV). Devices from the latest generation are equipped with Neural Processing Units (NPUs), which are specialized hardware architectures for running neural network-based algorithms commonly used in CV. However, state-of-the-art CV models are highly computationally demanding and remain unsuitable for NPUs and other edge processors optimized for efficiency and low power consumption.

Transmitting video to a remote device (e.g., a powerful server) allows the offloading of computation from the edge device. This processing paradigm is referred to as remote inference. However, it suffers from two major drawbacks: \textbf{1)} it places a heavy computational burden on the remote device and underutilizes the edge processor, and \textbf{2)} traditional video compression is not optimized for machine vision.

\begin{figure}[t]
    \centering
    \begin{minipage}[b]{1\linewidth}
    \centering
    \includegraphics[width=\textwidth]{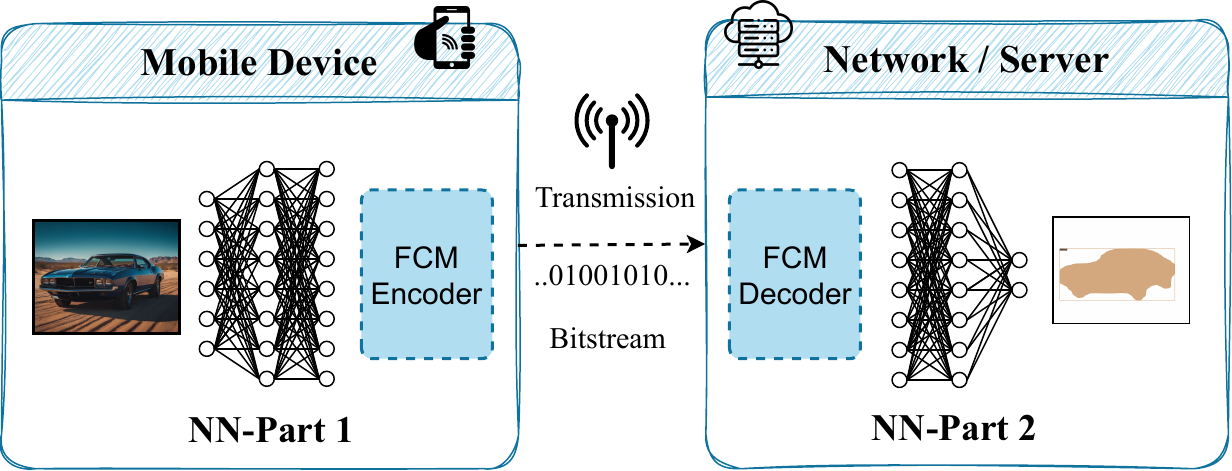}
    \end{minipage}
    \caption{An example of a split inference pipeline.}
\label{fig:split_inference}
\vspace{-0.6cm}
\end{figure}

Split inference~\cite{hyomin_vcm} is an emerging paradigm that addresses these issues. A neural network is partitioned into two components, NN Part-1 and NN Part-2, which are executed on separate devices, such as the edge device and the remote server, respectively. This process is illustrated in Fig.~\ref{fig:split_inference}. Video input is processed by NN Part-1, producing a stream of intermediate feature maps. These features are transmitted to the remote server, which completes inference using NN Part-2. 

While split inference addresses the challenge of resource utilization in remote inference, it does not resolve the inefficiency of traditional compression techniques. Intermediate features differ significantly from natural video, particularly in dimensionality and data distribution. To address this, the Moving Picture Experts Group (MPEG) introduced a standard for feature compression in split inference, known as Feature Coding for Machines (FCM)~\cite{fcm}. FCM tools enable the compression of features into a interoperable bitstream that is significantly smaller than raw intermediate data while preserving task accuracy. However, all tools introduced in VVC were originally designed for human-viewed video content. The rate-distortion optimization (RDO) in the VVC encoder targets perceptual metrics, such as PSNR, which are suboptimal for machine tasks. Furthermore, RDO may spend considerable time evaluating tools that are ultimately not selected, as they provide little benefit when compressing intermediate features. A more efficient encoder would proactively disable such tools, particularly when their usage is rare or counterproductive. This is especially advantageous for edge devices with limited computational resources. As such, we analyze VVC tool behavior under FCM and propose lightweight profiles optimized for feature compression in machine vision.

The remainder of the paper is organized as follows: Section~\ref{sec:fcm} provides a brief overview of FCM; Section~\ref{sec:experimental_results} analyzes VVC tool behavior and introduces three essential lightweight profiles; and finally, Section~\ref{sec:conclusion} summarizes the findings and concludes the paper.

\section{Feature Coding for Machines}
\label{sec:fcm}

\begin{figure}[t]
    \centering
    \begin{minipage}[b]{1\linewidth}
    \centering
    \includegraphics[width=\textwidth]{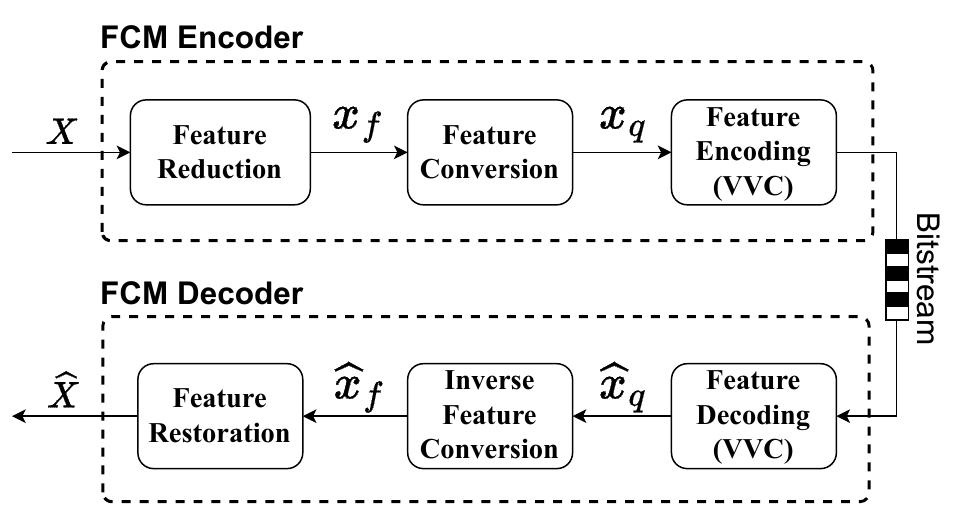}
    \end{minipage}
    \caption{A brief overview of FCM codec pipeline.}
\label{fig:fcm_overview}
\vspace{-0.4cm}
\end{figure}

\begin{figure}[t]
    \centering
    \begin{minipage}[b]{1\linewidth}
    \centering
    \includegraphics[width=\textwidth]{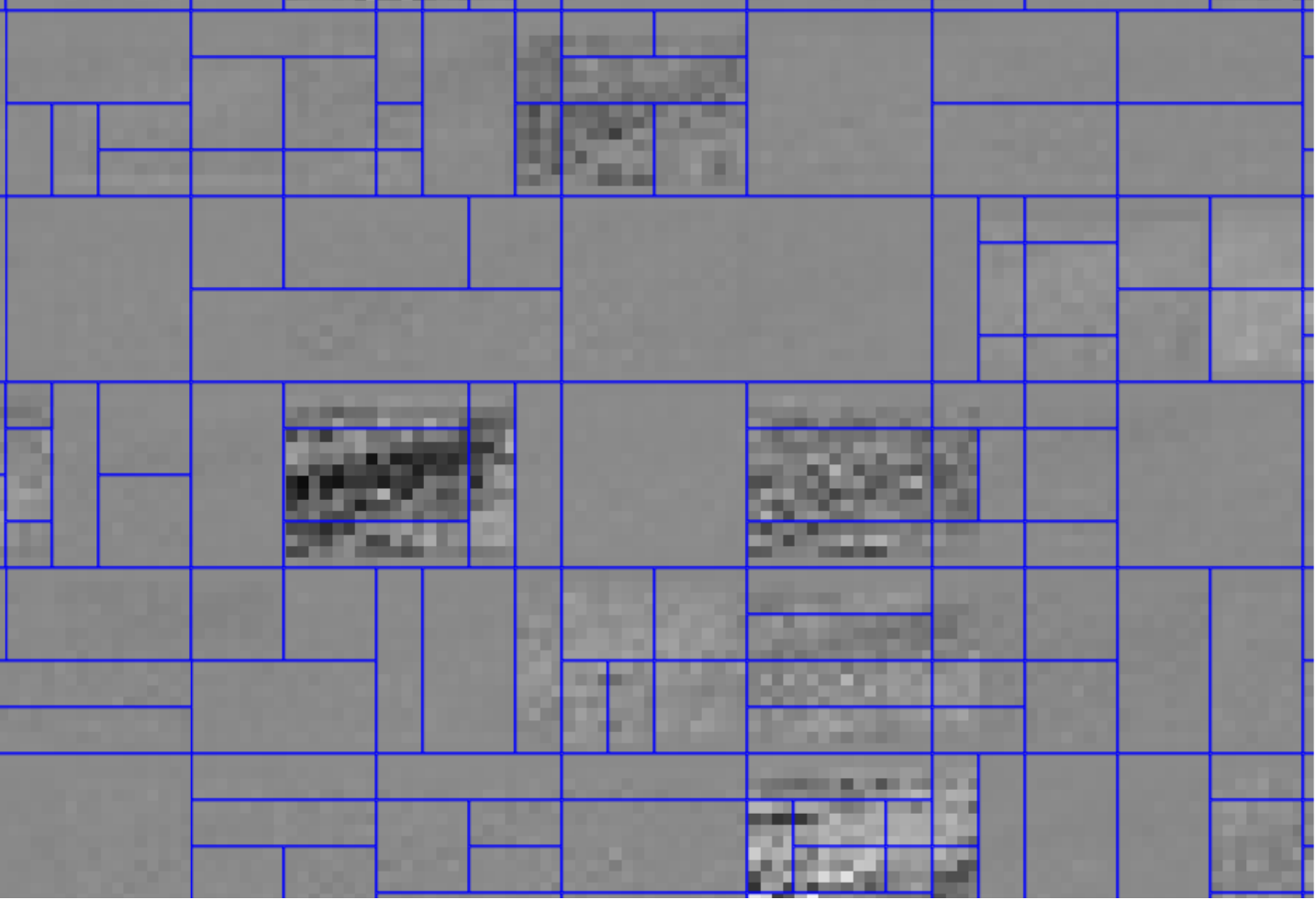}
    \end{minipage}
    \caption{Block partitioning visualization for a cropped region of the first frame in the \textit{Traffic} sequence, encoded at QP~19 using the Low-Delay configuration.}
    \vspace{-0.5cm}
    \label{fig:fcm_block_partition}
\end{figure}

Fig.~\ref{fig:fcm_overview} outlines the encoding and decoding process for intermediate features computed by a neural network partitioned into NN Part-1 and NN Part-2. $X$ represents the set of features tensors computed by NN Part-1, and $\hat{X}$ represents a lossy variant received by NN Part-2. Formally, $X=\{x_n\}_{n=1}^N$ is a set of $N$ feature tensors computed from an input video, where $x_n\in\mathbb{R}^{T \times C_n \times H_n \times W_n}$ is a 4-D feature tensor computed from $T$ input frames and has a size of $C_n$ channels, $H_n$ height, and $W_n$ width. The first stage of the FCM encoder reduces $X$ into a single fused feature tensor $x_f\in\mathbb{R}^{T_f \times C_f \times H_f \times W_f}$ with lower dimensions than all other feature tensors. The second stage converts the tensor into $x_q$, a video in which each frame is a 2-D image with all feature channels tiled, packed, and converted to 10-bit unsigned integers. The video is encoded with a standard video encoder such as VVC~\cite{vvc}. An example of a feature frame block partitioning by VVC encoder is shown in Fig.~\ref{fig:fcm_block_partition}. Finally, the FCM decoder takes the bitstream and performs the inverse of each encoder stage to produce $\hat{X}$.

\section{Low-complexity VVC Profiles}
\label{sec:experimental_results}

\begin{figure}[t]
    \centering
    \begin{minipage}[b]{1\linewidth}
    \centering
    \includegraphics[width=\textwidth]{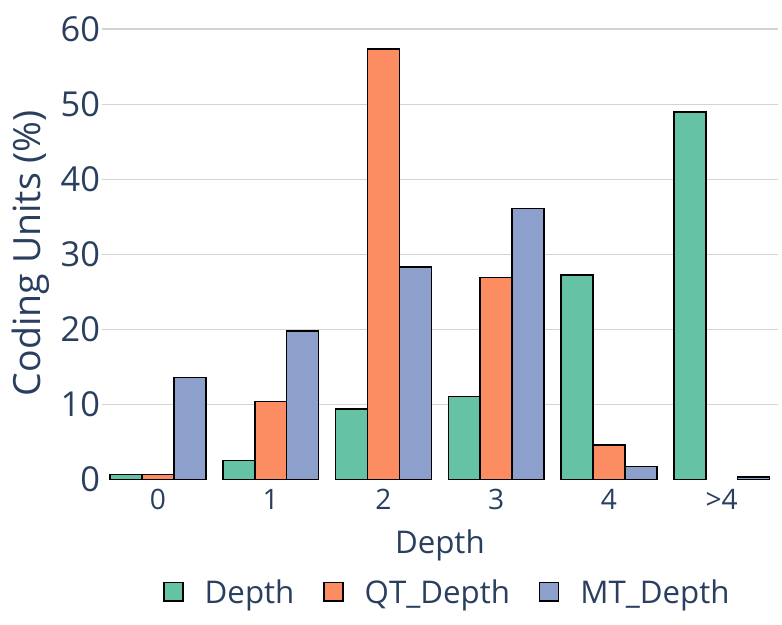}
    \end{minipage}
    \caption{Partition depth analysis. \texttt{Depth} denotes final CU depth; \texttt{QT\_Depth} and \texttt{MT\_Depth} correspond to quad-tree and multi-type tree split depths, respectively.}
\label{fig:depth_analysis}
\end{figure}

We evaluate VVC/H.266 for compressing intermediate features extracted from neural networks, with a focus on identifying a lightweight yet effective tool-set for Feature Coding for Machines (FCM). Unlike natural video content, these features exhibit sparse, abstract activation patterns, rendering many VVC tools—originally designed for perceptual quality—suboptimal for feature compression in split inference.

To identify essential tools, we begin by profiling the coding decisions made during encoding. Tools consistently skipped by the rate-distortion optimization (RDO) process are marked as candidates for removal. We then conduct a series of ablation experiments, where functionally related tool groups—particularly those newly introduced in VVC beyond HEVC/H.265~\cite{hevc}—are selectively disabled. This allows us to quantify their impact on both coding efficiency and downstream task performance, thereby guiding the construction of a low-complexity essential tool-set optimized for FCM.

\subsection{Evaluation setup}
\label{sec:experimental_setup}

We follow the Common Test and Training Conditions (CTTC) defined by MPEG~\cite{fcm_cttc} to evaluate feature compression for machine vision tasks. All experiments are conducted using the FCM Test Model (FCTM) v7.0\footnote{\url{https://git.mpeg.expert/MPEG/Video/fcm/fctm}}, a standardized reference implementation. For benchmarking, we adopt CompressAI-Vision~\cite{compressai_vision}, the official evaluation framework.

The evaluation consists of three datasets: SFU-HW, TVD, and HiEve. Each dataset is paired with a task-specific model and a split point, as summarized in Table~\ref{tbl:cttc}. Feature compression is performed using the VVC Test Model (VTM) v23.3~\cite{vtm}, configured in Low-Delay B mode.

For object detection, we use Faster RCNN-X101-FPN~\cite{faster_rcnn} for extracting features from the Feature Pyramid Network (FPN) layers. This yields four feature maps $X = \{ \mathbf{x}_n \}_{n=1}^4$, each with 256 channels and spatial resolution $H_n \times W_n = \frac{H_r}{2^{n+1}} \times \frac{W_r}{2^{n+1}}$, where $H_r \times W_r$ denotes the resized input.

For object tracking, we employ the JDE model~\cite{wang2019towards}, which is based on YOLOv3~\cite{EEERedmon2018_yolov3}, and evaluate two feature split strategies. The first strategy extracts Darknet backbone features: $\mathbf{x}_1 \in \mathbb{R}^{256 \times 76 \times 136}$, $\mathbf{x}_2 \in \mathbb{R}^{512 \times 38 \times 68}$, and $\mathbf{x}_3 \in \mathbb{R}^{1024 \times 19 \times 34}$. The second strategy uses reduced-channel variants with dimensions $128 \times 76 \times 136$, $256 \times 38 \times 68$, and $512 \times 19 \times 34$.

\begin{figure}[t]
    \centering
    \begin{minipage}[b]{1\linewidth}
    \centering
    \includegraphics[width=\textwidth]{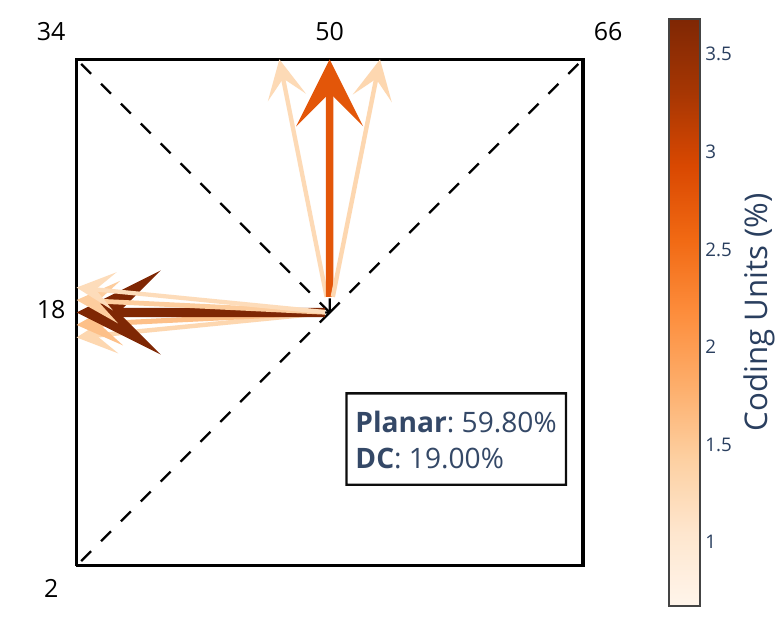}
    \end{minipage}
    \caption{ Intra prediction mode usage analysis. Arrow size and color indicate the percentage of CU using each mode (Mode 18: 3.68\%, Mode 50: 2.55\%)}
\label{fig:intra_modes)}
\end{figure}

\begin{figure*}[t]
    \centering
    \begin{minipage}[b]{1\linewidth}
    \centering
    \includegraphics[width=\textwidth]{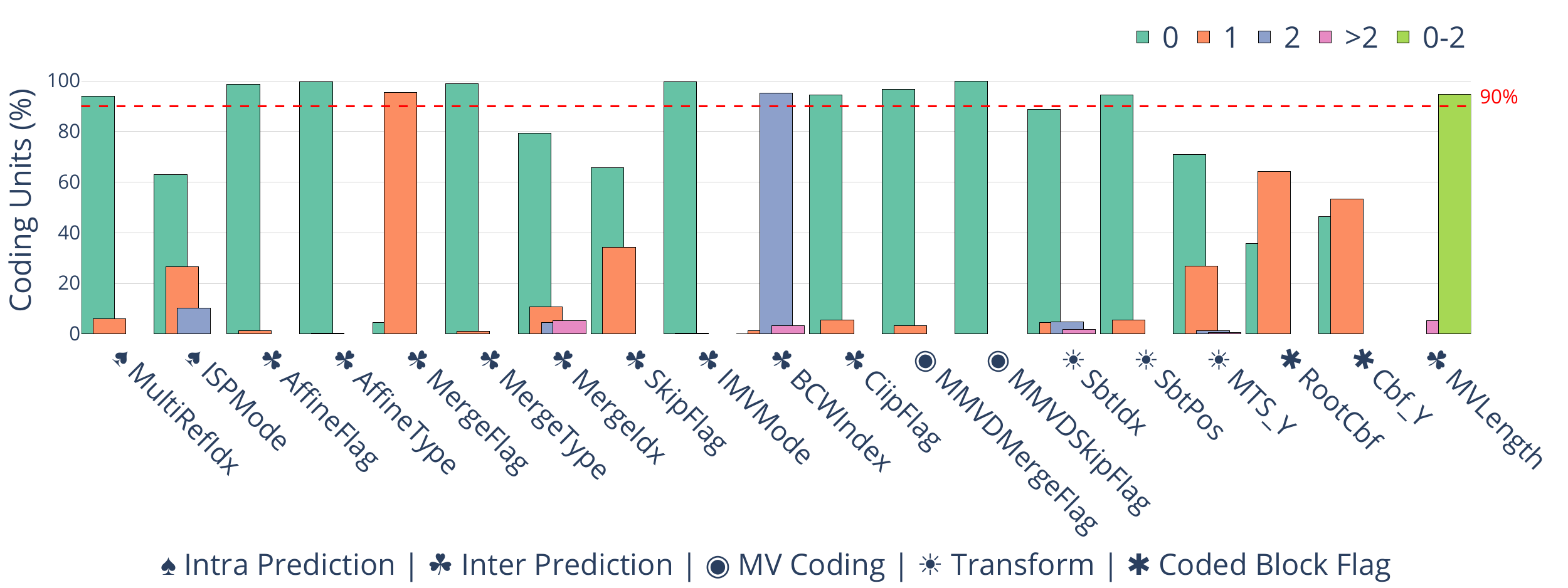}
    \end{minipage}
    \caption{Distribution of VVC coding modes selected by RDO for feature compression.}
\label{fig:mode_analysis)}
\end{figure*}

\subsection{Mode decision analysis}
\label{sec:mode_decidion_analysis}

As shown in Fig.~\ref{fig:depth_analysis}, the statistics of block partitioning-- \texttt{Depth}, \texttt{QT\_Depth}, and \texttt{MT\_Depth}—indicate active use of deep partitioning structures, with \texttt{MT\_Depth} peaking at depths $2$–$3$.  Such behavior is expected, given the irregular activation patterns typically found in intermediate feature maps, as illustrated in Fig.~\ref{fig:fcm_block_partition}.

Luma intra prediction statistics, as shown in Fig.~\ref{fig:intra_modes)}, show a skewed distribution. Of the 67 intra modes in VVC, Planar (mode $0$), DC ($1$), Horizontal ($18$), and Vertical ($50$) dominate, being used in over $85\%$ of all intra-coded blocks. While less frequent, the remaining angular modes are still needed to capture fine structures and maintain spatial alignment.

Among new intra tools in VVC, 
\texttt{MultiRefIdx} (multi-reference line index), which enables Multi-Reference Line (MRL) prediction, is rarely activated, as shown in Fig~\ref{fig:mode_analysis)}. In contrast, \texttt{ISPMode} (intra sub-partitioning mode) is often selected. All three values—$0$ (none), $1$ (vertical), and $2$ (horizontal)—are used, with vertical being most common than the horizontal.

For inter prediction, \texttt{MergeFlag} (merge mode) and \texttt{SkipFlag} (skip mode) are frequently activated. This is expected, as the 3D feature tensor is flattened into a 2D frame by stacking along the channel dimension, resulting in uniform motion across spatial locations in all channels. \texttt{MergeType} (merge type) is dominated by type $0$ (regular merge), and \texttt{MergeIdx} (merge index) prefers index $0$, showing spatial locality in motion reuse. \texttt{BCWIndex} (bi-prediction coding weight) peaks at index $2$, implying asymmetric weighting (e.g., $0.75$ forward, $0.25$ backward) is heavily used.

In contrast, several advanced VVC tools—\texttt{IMVMode} (integer motion vector mode), \texttt{AffineFlag} (affine motion), \texttt{AffineType} (affine model type), and \texttt{CiipFlag} (combined intra-inter prediction)—are rarely triggered. Likewise, \texttt{MMVDMergeFlag} and \texttt{MMVDSkipFlag}, which support Merge Mode with Motion Vector Difference (MMVD), show minimal usage. This is consistent with the \texttt{MVLength} (motion vector length) histogram, as shown in Fig.~\ref{fig:intra_modes)}, where most vectors fall within the $0$–$2$ range. These observations suggest that fine-grained motion refinement offers little advantage, and a reduced motion search range may suffice for feature encoding.

Transform tools, however, remain essential. Sub-block tools like \texttt{SbtIdx} (sub-block transform index), \texttt{SbtPos} (sub-block position), and \texttt{MTS\_Y} (multiple transform selection for luma) are frequently used, might be beneficial for spatial-frequency adaptation. Additionally, \texttt{RootCbf} and \texttt{Cbf\_Y} are frequently active, suggesting that over half of the blocks carry non-zero residuals.


\begin{table}[t]
\centering
\caption{Common Test \& Training Conditions (CTTC)~\cite{fcm_cttc}}
\vspace{-0.2cm}
\label{tbl:cttc}
\medskip
\resizebox{1.0\linewidth}{!}{%
\begin{tabular}{@{}c|c|c|c@{}}
\toprule
Dataset & Task & Network & Split Point (SP) \\ \midrule

SFU-HW~\cite{sfu_v1} 
& \begin{tabular}[c]{@{}c@{}}Object\\ Detection\end{tabular} 
& Faster RCNN-X101-FPN~\cite{faster_rcnn} 
& \begin{tabular}[c]{@{}c@{}} 4 SPs at\\ FPN \end{tabular} \\ \midrule

TVD~\cite{tvd} 
& \multirow{3}{*}{\begin{tabular}[c]{@{}c@{}}Object\\ Tracking\end{tabular}} 
& \multirow{3}{*}{JDE w/ YOLOv3~\cite{wang2019towards}}
& \begin{tabular}[c]{@{}c@{}}3 SPs at\\ Darknet-53~\cite{EEERedmon2018_yolov3}\end{tabular} \\ \cmidrule(l){1-1}\cmidrule(l){4-4}

HiEve~\cite{hieve} 
& 
& 
& \begin{tabular}[c]{@{}c@{}}3 SPs near\\ YOLO layers\end{tabular} \\

\bottomrule
\end{tabular}}
\vspace{-0.3cm}
\end{table}

\begin{table*}[t]
\centering
\caption{Comparison of coding efficiency and complexity across selected VVC configurations for feature compression. Results are reported in terms of BD-Rate and relative encoding/decoding time.}

\label{tbl:tool_impact}
\small
\begin{tabular}{@{}c>{\raggedright\arraybackslash}p{4.5cm}rrrrrrccc@{}}
\toprule
\multirow{3}{*}{\textbf{\#}} & \multirow{3}{*}{\textbf{\makecell{Configuration}}} &
\multicolumn{3}{c}{\textbf{Object Detection (BD-Rate)}} & 
\multicolumn{3}{c}{\textbf{Object Tracking (BD-Rate)}} &
\multirow{2}{*}{\textbf{ \makecell{Avg.\\BD-Rate}}} &
\multirow{2}{*}{\textbf{ \makecell{Enc.\\Time}}} &
\multirow{2}{*}{\textbf{ \makecell{Dec.\\Time}}}  \\
\cmidrule(lr){3-5} \cmidrule(lr){6-8}
& & SFU-A/B & SFU-C & SFU-D & TVD & \makecell{HiEve\\(1080p)} & \makecell{HiEve\\(720p)} & (\%)  & (\%) & (\%) \\
\midrule
\textbf{\textcolor{blue}{\Large\ding{172}}}   & Intra Modes $\rightarrow$ \{0, 1, 18, 50\}  & 9.83 & 7.23 & 6.25 & -0.57 & 1.48 & -0.10 & 4.02 & 93.11 & 99.35 \\
\textbf{\textcolor{teal}{\Large\ding{173}}}   & MaxMTTHierarchyDepth $\rightarrow$ 1 & -3.57 & 5.98 & -3.11 & 2.25 & 4.76 & 0.68 & 1.16 & 37.03 & 99.75 \\
\textbf{\textcolor{orange}{\Large\ding{174}}} & (Affine + SbTMVP) $\rightarrow$ 0 & -3.05 & 4.03 & 1.65 & -1.21 & 0.21 & -3.01 & \textbf{-0.23} & 84.29 & 99.39 \\
\textbf{\textcolor{violet}{\Large\ding{175}}} & (MTS + SbT + DepQuant) $\rightarrow$ 0 & 9.03 & 7.10 & 1.67 & 3.03 & 4.40 & 0.63 & 4.31 & 85.52 & 98.82 \\
\textbf{\textcolor{olive}{\Large\ding{176}}}  & (BCW + GEO + CIIP) $\rightarrow$ 0 & -2.75 & 5.33 & 5.55 & -0.61 & -0.65 & -1.39 & 0.91 & 90.50 & 98.54 \\
\textbf{\textcolor{magenta}{\Large\ding{177}}}& (SAO + DBF + ALF) $\rightarrow$ 0 & -14.89 & -5.60 & 9.32 & -4.00 & -0.19 & -2.38 & \textbf{-2.96} & 78.22 & 85.51 \\
\textbf{\textcolor{brown}{\Large\ding{178}}}  & (MRL + ISP) $\rightarrow$ 0 & -1.40 & 8.89 & 13.28 & -2.37 & 1.05 & 1.06 & 3.42 & 97.87 & 96.31 \\
\textbf{\textcolor{red}{\Large\ding{179}}}    & (MRL + Affine + IMV + CIIP + MMVD) $\rightarrow$ 0 \& \newline Motion Search $\rightarrow$ 16$\times$16 & -2.18 & 0.79 & 0.05 & 1.53 & 2.90 & 0.25 & 0.56 & 71.20 & 97.80 \\
\midrule
A & \textbf{\Large\textcolor{orange}{\ding{174}}+\textcolor{magenta}{\ding{177}}} & -12.34 & 0.63 & 1.81 & 1.65 & 0.79 & -1.78 & \textbf{-1.54} & 66.93 & 91.50 \\
B & \textbf{\Large\textcolor{orange}{\ding{174}}+\textcolor{magenta}{\ding{177}}+\textcolor{red}{\ding{179}}} & -12.76 & -0.10 & 20.12 & 10.48 & 3.48 & -2.82 & 3.07 & 50.56 & 89.15 \\
C & \textbf{\Large\textcolor{orange}{\ding{174}}+\textcolor{magenta}{\ding{177}}+\textcolor{red}{\ding{179}}+\textcolor{olive}{\ding{176}}} & -23.39 & 5.34 & 11.83 & -3.43 & 0.44 & -1.87 & \textbf{-1.85} & 48.54 & 90.97 \\
D & \textbf{\Large\textcolor{orange}{\ding{174}}+\textcolor{magenta}{\ding{177}}+\textcolor{red}{\ding{179}}+\textcolor{olive}{\ding{176}}+\textcolor{teal}{\ding{173}}} & -11.68 & -0.34 & 16.84 & 11.84 & 4.51 & 1.71 & 3.81 & 5.55 & 86.59 \\
E & \textbf{\Large\textcolor{orange}{\ding{174}}+\textcolor{magenta}{\ding{177}}+\textcolor{red}{\ding{179}}+\textcolor{olive}{\ding{176}}+\textcolor{teal}{\ding{173}}+\textcolor{brown}{\ding{178}}} & -5.54 & -1.61 & 6.09 & 10.89 & 1.40 & -0.98 & 1.71 & 4.34 & 86.62 \\
F & \textbf{\Large\textcolor{orange}{\ding{174}}+\textcolor{magenta}{\ding{177}}+\textcolor{red}{\ding{179}}+\textcolor{olive}{\ding{176}}+\textcolor{teal}{\ding{173}}+\textcolor{brown}{\ding{178}}+\textcolor{blue}{\ding{172}}} & -3.56 & 5.05 & 4.68 & 10.23 & 4.05 & -2.59 & 2.98 & 4.05 & 85.39 \\
G & \textbf{\Large\textcolor{orange}{\ding{174}}+\textcolor{magenta}{\ding{177}}+\textcolor{red}{\ding{179}}+\textcolor{olive}{\ding{176}}+\textcolor{teal}{\ding{173}}+\textcolor{brown}{\ding{178}}+\textcolor{blue}{\ding{172}}+\textcolor{violet}{\ding{175}}} & 0.06 & 8.25 & 3.14 & 9.92 & 4.01 & 0.20 & 4.26 & 4.51 & 84.65 \\
\bottomrule
\end{tabular}

\vspace{0.8em}
\noindent
\textbf{\textcolor{blue}{\Large\ding{172}}} Intra prediction \quad
\textbf{\textcolor{teal}{\Large\ding{173}}} Block partitioning \quad
\textbf{\textcolor{orange}{\Large\ding{174}}} Sub-block motion compensation \quad
\textbf{\textcolor{violet}{\Large\ding{175}}} Transform tools \quad
\textbf{\textcolor{olive}{\Large\ding{176}}} Motion compensation \quad\\
\textbf{\textcolor{magenta}{\Large\ding{177}}} In-loop filtering \quad
\textbf{\textcolor{brown}{\Large\ding{178}}} Intra tools \quad
\textbf{\textcolor{red}{\Large\ding{179}}} Tools frequently skipped by RDO (used in fewer than 10\% coding units)

\end{table*}

\subsection{Lightweight essential tool-set}
\label{ssec:tool_set}
Building on the mode decision analysis, we evaluate the impact of selectively disabling functionally grouped VVC tools. Table~\ref{tbl:tool_impact} summarizes the trade-offs across eight configurations in terms of rate-distortion performance and computational complexity. Each configuration targets a specific functional category of tools. All results are reported relative to the default VVC setup used for Feature Coding for Machines (FCM)~\cite{fcm_cttc}, and include BD-Rate~\cite{bd_rate}, encoding time, and decoding time.

Configuration \symb{blue}{172} limits luma intra prediction to just four commonly used modes—Planar, DC, Horizontal, and Vertical. While this reduces complexity, it increases BD-Rate by $4.02\%$, showing that even the less frequently used angular modes still help capture important spatial details. Configuration \symb{teal}{173} restricts partitioning to just one level of MTT, leading to a large $63\%$ reduction in encoding time with only a $1.16\%$ increase in BD-Rate.

The next group of configurations—\symb{orange}{174} through \symb{brown}{178}—targets tools newly introduced intra and inter tool in VVC to improve visual quality for human viewers and generally involve more advanced coding strategies. Configuration \symb{orange}{174} disables sub-block motion compensation tools, specifically affine motion and Sub-block Temporal Motion Vector Prediction (SbTMVP), and improves BD-Rate by $-0.23\%$, suggesting that these complex motion models offer limited benefit for feature compression. Configuration \symb{violet}{175} turns off transform-related tools, including MTS, Sub-Block Transform (SBT), and Dependent Quantization (DepQuant). This leads to a $4.31\%$ increase in BD-Rate, highlighting their importance in capturing frequency-domain structure in features. Configuration \symb{olive}{176} disables advanced motion compensation tools such as BCW, Geometric Partitioning Mode (GEO), and CIIP. Disabling this group of tools results in a $0.91\%$ BD-Rate increase.

In-loop filters—Sample Adaptive Offset (SAO), Deblocking Filter (DBF), and Adaptive Loop Filter (ALF)—are evaluated in configuration \symb{magenta}{177}. While they are important for visual quality in traditional video coding, we find they negatively impact feature compression for machines. Disabling them improves BD-Rate by $-2.96\%$ and significantly reduces complexity, suggesting they distort activation values used by the downstream task network. Configuration \symb{brown}{178} disables both \texttt{MultiRefIdx} and \texttt{ISP}. Since \texttt{ISPMode} is frequently used and contributes to better directionality, turning it off leads to a $3.42\%$ BD-Rate increase. Finally, configuration \symb{red}{179} combines the least-used tools inspired by our mode analysis—\texttt{MultiRefIdx}, \texttt{Affine}, \texttt{IMV}, \texttt{CIIP}, and \texttt{MMVD}—and limits motion search range to $16$ pixels. Despite simplifying several parts of the codec, it only increases BD-Rate by $0.56\%$ while reducing encoding time by $28.8\%$.

We then explore a series of incremental combinations--
configurations A through G--that progressively disable more tools in order of increasing BD-Rate loss. Configurations \symb{orange}{174} and \symb{magenta}{177} are considered effective entry points, offering favorable trade-offs between encoding complexity and BD-Rate. 

\begin{figure}[H]
    \centering
    \begin{minipage}[b]{1\linewidth}
    \centering
    \includegraphics[width=\textwidth]{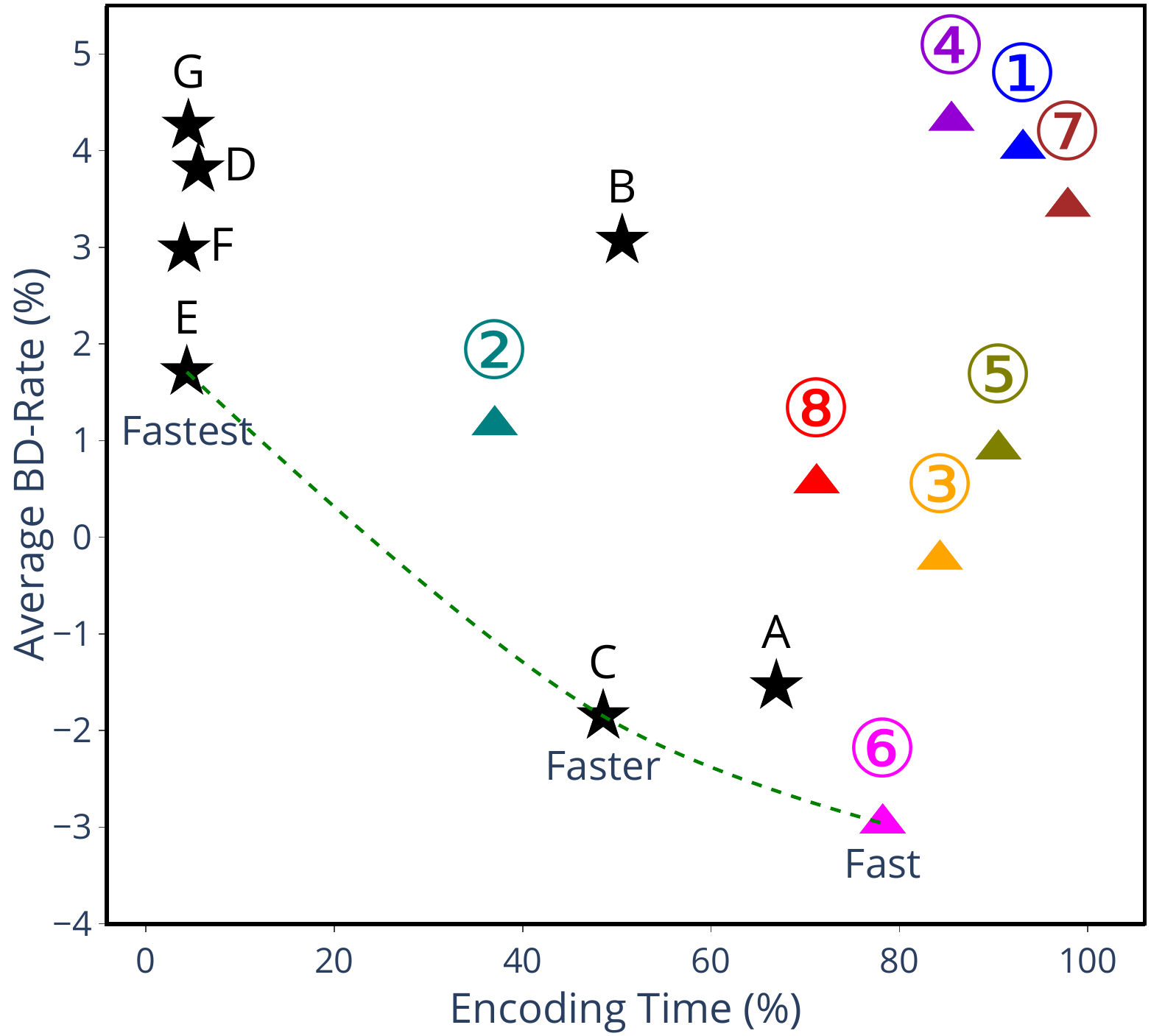}
    \end{minipage}
    \caption{Average BD-Rate versus encoding time trade-offs for all the configurations in Table~\ref{tbl:tool_impact}. Triangle markers ($\triangle$) denote individual tool groups (\symb{blue}{172}–\symb{red}{179}), and star markers ($\star$) represent combined configurations (A–G). The proposed \textit{Fast}, \textit{Faster}, and \textit{Fastest} profiles are annotated.}
    
\label{fig:profile)}
\end{figure}

To identify optimal low-complexity toolsets, we evaluate all configurations and visualize the rate-distortion versus complexity trade-offs in Fig.~\ref{fig:profile)}. Based on this analysis, we highlight three practical configurations: \textbf{Fast} (\symb{magenta}{115}\symb{magenta}{177}), which achieves a $21.8\%$ reduction in encoding time while improving BD-Rate by $-2.96\%$; \textbf{Faster} (\symb{black}{72}C), which reduces encoding time by $51.5\%$ with a $-1.85\%$ BD-Rate gain; and \textbf{Fastest} (\symb{black}{72}E), which provides the highest speedup—$95.6\%$ reduction in encoding time—at the cost of a $1.71\%$ increase in BD-Rate.

Among all configurations, we found that three i.e.,\symb{violet}{175}, \symb{blue}{172}, \symb{brown}{178} are particularly important for feature compression for machines, as shown in Fig.~\ref{fig:profile)}. Disabling transform tools (\symb{violet}{175}) results in the most significant performance degradation, as these tools are essential for capturing fine-grained frequency components within feature maps. Limiting luma intra prediction to only four modes (\symb{blue}{172}) disrupts spatial alignment and reduces the model’s ability to reconstruct structured feature activations. Additionally, removing intra tools such as ISP (\symb{brown}{178}), which is frequently selected during mode decision, leads to performance loss. In contrast, retaining in-loop filters (\symb{magenta}{177}), although beneficial for perceptual quality, negatively affects the statistical distribution of activations, thereby impairing downstream task performance and should always be disabled for feature compression.


\section{Conclusion}
\label{sec:conclusion}
We present a comprehensive analysis of VVC coding tools for compressing intermediate features in split-inference systems. By profiling encoder decisions and conducting targeted ablation studies, we identify a subset of tools—including multiple transform selection (MTS), sub-block transforms (SbT), dependent quantization (DepQuant), intra sub-partitioning (ISP), and directional intra modes—as essential for preserving the spatial and frequency structure of feature activations. In contrast, tools such as affine motion, integer motion vectors (IMV), combined intra-inter prediction (CIIP), multi-reference line (MRL), and merge mode with motion vector difference (MMVD) are frequently skipped by rate-distortion optimization and can be disabled without compromising task performance. Notably, disabling in-loop filters yields a $2.96\%$ average BD-Rate improvement and up to $14.89\%$ for object detection. Based on these insights, we propose three profiles—Fast, Faster, and Fastest. The proposed Fastest profile reduces encoding time by $95.6\%$ with only a $1.71\%$ BD-Rate loss, suggesting that effective feature compression requires a tool-set distinct from those optimized for human visual perception.

\bibliographystyle{IEEEbib}
\bibliography{strings,refs}

\end{document}